\pdfoutput=1
\documentclass{article}

\PassOptionsToPackage{numbers, compress}{natbib}
\usepackage[preprint]{neurips_2026}

\usepackage{amsmath,amssymb,amsfonts,mathtools}
\usepackage{nicefrac}
\DeclareMathOperator{\clip}{clip}

\usepackage{graphicx}
\usepackage{caption}
\usepackage{subcaption}
\usepackage{wrapfig}
\usepackage{multirow}
\usepackage{booktabs}
\usepackage{placeins}
\usepackage{needspace}
\usepackage{microtype}

\usepackage[utf8]{inputenc}
\usepackage[T1]{fontenc}
\DeclareUnicodeCharacter{221A}{\ensuremath{\surd}}

\usepackage[table]{xcolor}
\definecolor{pos}{HTML}{2E7D32}
\definecolor{neg}{HTML}{C62828}

\usepackage[most]{tcolorbox}

\usepackage{url}
\usepackage[colorlinks,
            linkcolor=black,
            anchorcolor=blue,
            citecolor=teal]{hyperref}

\title{Resolving Action Bottleneck: Agentic Reinforcement Learning Informed by Token-Level Energy}

\author{\textbf{Langzhou He}\textsuperscript{1},
\textbf{Junyou Zhu}\textsuperscript{2,3},
\textbf{Yue Zhou}\textsuperscript{1},
\textbf{Zhengyao Gu}\textsuperscript{1},\\
\textbf{Junhua Liu}\textsuperscript{4},
\textbf{Wei-Chieh Huang}\textsuperscript{1},
\textbf{Henry Peng Zou}\textsuperscript{1},
\textbf{David Wipf}\textsuperscript{5},\\
\textbf{Philip S. Yu}\textsuperscript{1},
\textbf{Qitian Wu}\textsuperscript{6}\thanks{Corresponding author.}\\[0.5em]
\textsuperscript{1}University of Illinois Chicago \quad
\textsuperscript{2}Potsdam Institute for Climate Impact Research \\
\textsuperscript{3}Technical University of Berlin \quad
\textsuperscript{4}University of Southern California \\
\textsuperscript{5}University of Hong Kong \quad
\textsuperscript{6}Broad Institute of MIT and Harvard \\[0.5em]
\texttt{\{lhe24,yzhou232,zgu24,whuang80,pzou3,psyu\}@uic.edu} \\
\texttt{junyou.zhu@pik-potsdam.de} \quad \texttt{jliu2321@usc.edu} \\
\texttt{dwipf@hku.hk} \quad \texttt{wuqitian@broadinstitute.org}
}

\begin{document}

\maketitle

\begin{abstract}
Agentic reinforcement learning trains large language models using multi-turn trajectories that interleave long reasoning traces with short environment-facing actions. Common policy-gradient methods, such as PPO and GRPO, treat each token in a trajectory equally, leading to uniform credit assignment. In this paper, we critically demonstrate that such uniform credit assignment largely misallocates token-level training signals. From an energy-based modeling perspective, we show that token-level training signals, quantified by their correlations with reward variance of different rollouts sampled from a given prompt, concentrate sharply on action tokens rather than reasoning tokens, even though action tokens account for only a small fraction of the trajectory. We refer to this phenomenon as the Action Bottleneck. Motivated by this observation, we propose an embarrassingly simple token reweighting approach, \textsc{ActFocus}, that downweights gradients on reasoning tokens, along with an additional energy-based redistribution mechanism that further increases the weights on action tokens with higher uncertainty. Across four environments and different model sizes, \textsc{ActFocus} consistently outperforms PPO and GRPO, yielding final-step gains of up to 65.2 and 63.7 percentage points, respectively, without any additional runtime or memory cost.
\end{abstract}

\section{Introduction}

As large language models (LLMs) are increasingly deployed as autonomous agents that combine reasoning~\citep{guo2025deepseek, jaech2024openai}, planning~\citep{zhou2023webarena}, and tool use~\citep{wolflein2025llmagentsmakingagent} to solve multi-turn, long-horizon tasks, reinforcement learning (RL) has emerged as the natural post-training framework for sequential decision-making settings~\citep{wang2025ragenunderstandingselfevolutionllm, wang2024survey, xi2025agentgymrltrainingllmagents, abdulhai2023lmrlgymbenchmarksmultiturn}. Despite rapid progress with different policy optimization methods~\citep{schulman2017proximalpolicyoptimizationalgorithms, shao2024deepseekmathpushinglimitsmathematical}, training LLM agents with RL remains difficult. One of the core challenges is credit assignment~\citep{pignatelli2024surveytemporalcreditassignment, kazemnejad2025vinepporefiningcreditassignment}, where, in multi-turn trajectories, a sparse episodic reward must be propagated back through thousands of generated tokens, making it difficult to determine which parts of a trajectory should drive learning. The problem is compounded by optimization instability~\citep{cui2025entropymechanismreinforcementlearning}, as delayed feedback, long generated traces, and noisy policy updates often lead to degraded late-stage performance or collapse. 

Recent evidence shows that multi-turn, underspecified conversations substantially increase outcome variability for LLMs, sharpening the need for reliable credit assignment in agentic RL \citep{laban2025llmslostmultiturnconversation}. In a typical agentic RL rollout, such outcome variability often hinges on a few critical environment-facing action tokens that drive state transitions, yet these tokens are buried within massive amounts of reasoning tokens in the trajectory \citep{yao2023reactsynergizingreasoningacting, schick2023toolformerlanguagemodelsteach, wen2024reinforcinglanguageagentspolicy}. Even though these action tokens are the only parts of the output that directly interact with the environment, standard policy optimization methods treat all generated tokens uniformly \citep{schulman2017proximalpolicyoptimizationalgorithms, liu2025uniformheterogeneoustailoringpolicy}. Consequently, gradient updates are spread equally across the whole trajectory, diluting training signals for the critical action tokens that trigger the decision-making~\citep{chan2024denserewardfreereinforcement, vassoyan-etal-2025-ignore}. In multi-turn settings, this mismatch compounds with each additional turn: reasoning tokens keep accumulating while the environment is driven by only a few informative actions. 

\begin{figure}[!t]
  \centering
  \includegraphics[width=\linewidth]{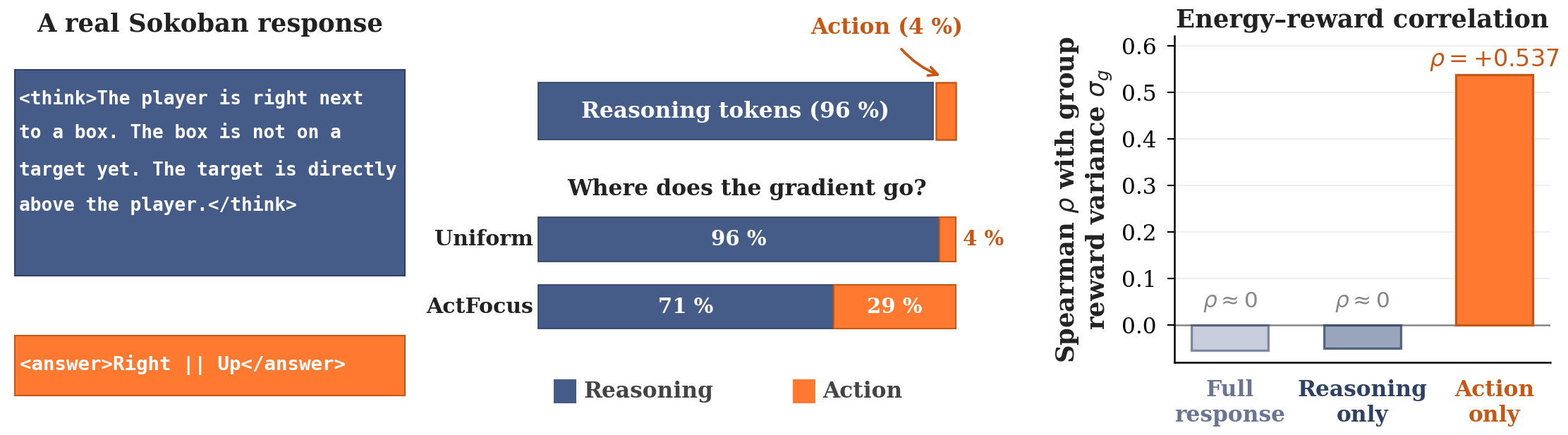}
  \caption{\textbf{Action Bottleneck in agentic reinforcement learning.}
\emph{Left:} A real response from a trajectory, where reasoning tokens far outnumber action tokens. 
\emph{Middle:} Action tokens constitute only 4\% of model-generated tokens; our token-level reweighting redirects gradient mass towards them. 
\emph{Right:} Training signal concentrates in action spans. Results are from Sokoban~3B.}
  \label{fig:action-bottleneck}
    \vspace{-10pt}
\end{figure}

Figure~\ref{fig:action-bottleneck} shows that this structural mismatch manifests as token-level signal sparsity in agentic RL. In a representative Sokoban trajectory, action tokens constitute only 4\% of the generated output, while reasoning tokens account for the remaining 96\%. Under uniform credit assignment, gradient mass therefore follows token count and is dominated by the reasoning majority. However, in this paper, we demonstrate that the training signals follow the opposite pattern. We quantify each token's training signal as its predictive uncertainty under a frozen reference model, measured by the free energy of its next-token predictive distribution.\footnote{An ideal indicator for quantifying training signals should reveal where outcome uncertainty concentrates, stay stable across training, and be computationally inexpensive. Energy, used in this work, satisfies all of these criteria; alternatives such as entropy and log-likelihood are compared empirically in Sec.~\ref{sec:signal-sweep}.} To further evaluate how informative the signal is for training, we hinge it on outcome variability: for each prompt, we calculate the variance of episodic rewards $\sigma_g$ across multiple rollouts (sampled from the same prompt), which captures how uncertain the current policy is about the prompt's outcome. We find that the mean energy of action tokens is strongly correlated with reward variance (with $\rho=+0.537$), while the mean energy of reasoning tokens or full trajectories, which average over both action and reasoning tokens, is nearly uninformative (with $\rho\approx 0$). \textbf{This suggests that informative training signals are concentrated in a small number of action tokens rather than being distributed uniformly across the trajectory.} We refer to this phenomenon as the \emph{Action Bottleneck}.

Motivated by this observation, we propose an energy-based approach to token-level credit assignment in agentic RL. The method is conceptually intuitive and can be decomposed into two aspects: 1) We downweight gradients on reasoning tokens, which is equivalent to increasing the weights on action tokens and helps to redirect gradient mass toward the sparse action span that determines environmental outcomes; 2) We utilize token-level energy to further redistribute the weights across action tokens, which prioritizes action tokens with high energy (i.e., more predictive uncertainty). In this way, we transform our empirical observations into an actionable approach that systematically guides the model to better leverage the informative training signals underlying a few critical tokens in the long trajectory.

We evaluate our approach, \textsc{ActFocus}, on four representative multi-turn agentic environments using Qwen2.5-Instruct \citep{qwen2025qwen25technicalreport} models at multiple scales, and train each configuration with both PPO \citep{schulman2017proximalpolicyoptimizationalgorithms} and GRPO \citep{shao2024deepseekmathpushinglimitsmathematical}. Through these experiments, we find that \textsc{ActFocus} produces consistent performance gains over PPO and GRPO, both of which use uniform credit assignment, with final success-rate improvements of up to 65.2 and 63.7 percentage points, respectively. Beyond absolute performance metrics, we observe that \textsc{ActFocus} effectively improves GRPO's training stability by substantially reducing peak-to-final degradation. Our main contributions are summarized as follows:

\textbf{(i)} We analyze token-level attributions in agentic RL training through an energy-based modeling perspective and reveal a counter-intuitive phenomenon: while reasoning tokens dominate in long-horizon multi-turn trajectories, infrequent action tokens contribute the majority of informative training signals. In particular, the uncertainty in action tokens, as measured by the free energy of next-token predictive distributions, is highly correlated with outcome variability. 

\textbf{(ii)} We propose a token reweighting mechanism that downweights gradients on reasoning tokens and additionally redistributes the weights on action tokens according to their uncertainty levels as characterized by the free energy. This simple scheme shifts gradient mass towards the most informative action tokens while remaining fully compatible with common policy-gradient algorithms.

\textbf{(iii)} Through extensive experiments across four environments and multiple model scales, \textsc{ActFocus} consistently outperforms PPO and GRPO, with average gains of 17.0 and 34.6 percentage points and final-step gains of up to 65.2 and 63.7 percentage points, respectively. In particular, \textsc{ActFocus} can significantly enhance the training stability of GRPO by mitigating the peak-to-final degradation observed under uniform credit assignment.

\section{Related Work}
\label{relatedwork}

\paragraph{Agentic Reinforcement Learning for LLMs.}
Reinforcement learning has become a central paradigm for aligning large language models with sequential decision-making objectives beyond single-turn text generation~\citep{ouyang2022traininglanguagemodelsfollow}. 
Recent work extends RL to multi-turn agent settings from different angles. For instance, LOOP~\citep{chen2025reinforcementlearninglonghorizoninteractive} formulates interactive digital agents as RL in stateful environments, while RAGEN~\citep{wang2025ragenunderstandingselfevolutionllm} presents a general framework for multi-turn agent training under stochastic environments. ARLArena~\citep{wang2026arlarenaunifiedframeworkstable} systematizes the study of training stability by decomposing policy-gradient design into core dimensions. However, despite these advances, credit assignment in existing agentic RL methods remains relatively under-explored and does not explicitly distinguish between different parts of an agent trajectory, particularly reasoning and action tokens.

\paragraph{Credit Assignment in Agentic RL.}
A growing line of work argues that uniform credit assignment largely under-represents the token-level training signals of long-horizon agent trajectories. Most existing efforts operate at the step or trajectory level. CARL~\citep{shen2026carlfocusingagenticreinforcement} uses policy entropy as a proxy for action importance and concentrates optimization on critical actions across turns. GiGPO~\citep{feng2025groupingrouppolicyoptimizationllm} introduces a two-level advantage estimator based on anchor states that reappear across trajectories, enabling step-level relative credit assignment. iStar~\citep{liu2025agenticreinforcementlearningimplicit} learns implicit step-wise rewards through trajectory-level preference optimization and combines step-level and trajectory-level signals in policy updates. These methods recognize that credit should vary across an agent trajectory, but they remain coarse in granularity and do not distinguish reasoning tokens from action tokens within each step. Several works in reasoning RL have explored finer-grained reward redistribution and intermediate supervision~\citep{tan2026gtpogrpostokensequencelevel,cui2025processreinforcementimplicitrewards, ma2026fipoelicitingdeepreasoning}, but these approaches are designed for single-turn reasoning without diving into the reasoning-action disentanglement in multi-turn agent trajectories.

\section{Preliminaries}
\label{sec:prelim}

\subsection{Multi-Turn Agentic RL}

We formulate multi-turn agentic RL as a Markov Decision Process $\mathcal{M} = \langle \mathcal{S}, \mathcal{A}, P, R \rangle$~\citep{sutton1998reinforcement}, where states $s \in \mathcal{S}$ are interaction histories, actions $a \in \mathcal{A}$ are environment-facing operations available to the agent, the transition $P(s_{k+1} \mid s_k, a_k)$ specifies how the environment responds to an action, and the reward $R(\tau)$ assigns a scalar episodic return based on the terminal state of the trajectory. At each turn $k \in \{1, \dots, K\}$, the policy $\pi_\theta$ observes the current state $s_k$ and generates a structured response:
\begin{equation}
\label{eq:response}
o_k = \texttt{<think>}\dots\texttt{</think>}\;\texttt{<answer>}\;a_k\;\texttt{</answer>},
\end{equation}
where the \texttt{<think>} block contains the agent's reasoning and the \texttt{<answer>} block contains the action $a_k$. The environment extracts $a_k$ from $o_k$, transitions to the next state $s_{k+1}$, and returns a reward $R(\tau)$ at episode termination. The full trajectory is $\tau = (s_1, o_1, s_2, o_2, \dots, s_K, o_K)$, and the RL objective is to maximize expected episodic reward: $J(\theta) = \mathbb{E}_{\tau \sim \pi_\theta}[R(\tau)]$.

\paragraph{Token-level policy optimization.}
Each response $o_k$ consists of tokens $(y_{k,1}, \dots, y_{k,T_k})$ from both the reasoning and action segments. For notational simplicity, we drop the turn index and use $y_t$ for a generic response token. Both PPO and GRPO optimize this objective via a clipped token-level surrogate loss that shares the same general form:
\begin{equation}
\label{eq:ppo-loss}
\mathcal{L}(\theta) =
-\frac{1}{|\mathcal{D}|}
\sum_{(y_t,\hat A_t)\in\mathcal{D}}
w_t \cdot
\min\!\Big(
\rho_t(\theta)\hat A_t,\;
\mathrm{clip}\big(\rho_t(\theta),1-\epsilon,1+\epsilon\big)\hat A_t
\Big),
\end{equation}
where $\rho_t(\theta) = \pi_\theta(y_t\mid c_t) / \pi_{\theta_{\mathrm{old}}}(y_t\mid c_t)$ is the importance ratio between the current and old policy, $\epsilon$ is the clipping threshold, and $\hat{A}_t$ is the per-token advantage. Here, $c_t$ denotes all tokens preceding $y_t$, to distinguish it from the environment state $s_k$, and $w_t$ is a per-token weight.

PPO estimates $\hat{A}_t$ via Generalized Advantage Estimation (GAE)~\citep{schulman2018highdimensionalcontinuouscontrolusing} with a learned value function, providing a distinct advantage for each token position. GRPO eliminates the critic and instead computes a single trajectory-level advantage:
\begin{equation}
\label{eq:grpo-advantage}
\hat{A}_\tau = \frac{R(\tau) - \mathrm{mean}(R(\tau))}{\mathrm{std}(R(\tau))},
\end{equation}
where the mean and standard deviation are computed over all trajectories sampled from the same prompt. Under GRPO, every token in a trajectory shares the same advantage: $\hat{A}_t = \hat{A}_\tau$ for all $t$ in $\tau$. In both algorithms, the common practice is to set $w_t = 1$ for all response tokens, which corresponds to the uniform credit assignment that we critically challenge in this paper. Complete descriptions for algorithm-specific objectives and implementation details are deferred to Appendix~\ref{app:rl-objectives}.

\begin{table}[t]
  \centering
  \caption{Action token fraction in inference-only rollouts before RL training. The imbalance is consistent across environments and model scales.}
  \label{tab:token-composition}
  \small
  \begin{tabular*}{\textwidth}{@{\extracolsep{\fill}}llccc@{}}
    \toprule
    \textbf{Env.} & \textbf{Scale} & \textbf{Tokens} & \textbf{Think} & \textbf{Act} \\
    \midrule
    Sokoban    & 0.5B  & 225   & 90.5\% & 9.5\%  \\
    Sokoban    & 3B    & 202   & 96.1\% & 3.9\%  \\
    FrozenLake & 0.5B  & 284   & 95.9\% & 4.1\%  \\
    FrozenLake & 3B    & 240   & 97.4\% & 2.6\%  \\
    Sudoku     & 0.5B  & 400   & 84.7\% & 15.3\% \\
    Sudoku     & 1.5B  & 397   & 85.8\% & 14.2\% \\
    WebShop    & 1.5B  & 1,951 & 89.2\% & 10.8\% \\
    WebShop    & 3B    & 2,133 & 92.7\% & 7.3\%  \\
    \bottomrule
  \end{tabular*}
\end{table}

\subsection{Frozen-Reference Token Energy}
\label{sec:prelim-energy}

Recent work~\citep{grathwohl2020classifiersecretlyenergybased, xie2016theory} establishes an equivalence between neural classifiers and energy-based models (EBMs). Given a classifier that maps inputs to $C$-dimensional logits $h(x) \in \mathbb{R}^C$, the classifier-EBM equivalence is established by setting the energy of an input-label pair as $E(x, y) = -h(x)[y]$. Marginalizing over labels yields a free energy $E(x) = -\log \sum_{c} \exp(h(x)[c])$, which captures the uncertainty of the model for a given input~\citep{liu2020energy}: lower energy indicates a confident, peaked prediction, while higher energy indicates uncertainty across multiple plausible outputs.

We extend this idea to autoregressive language models, which predict a distribution over the vocabulary at each position. At each token position $t$, a frozen reference model produces next-token logits $f_t^{\mathrm{ref}} \in \mathbb{R}^{|V|}$ over vocabulary $V$, conditioned on the preceding context. We define the \textbf{token-level energy} as:
\begin{equation}
\label{eq:token-energy}
E_t = -\log \sum_{v \in V} \exp\!\big(f_{t,v}^{\mathrm{ref}}\big).
\end{equation}
A lower $E_t$ corresponds to higher confidence under the reference model, while a higher $E_t$ indicates greater uncertainty about the next token. This signal is well-suited for diagnosing agentic RL for two reasons. First, because the reference model is frozen, $E_t$ remains stable throughout training, unlike policy-derived quantities such as entropy or confidence that drift as the policy evolves. Second, computing $E_t$ requires only a single forward pass, making it cheaper than Monte Carlo estimates that require repeated rollouts~\citep{shen2026carlfocusingagenticreinforcement}. We use token-level energy to analyze where the reward-predictive signal concentrates within agent trajectories, and this analysis motivates the intervention in Sec.~\ref{sec:method}.

\section{Token Credit Assignment}
\label{sec:method}
We now turn the Action Bottleneck diagnosis into a token-level weighting mechanism that aims to address two issues: 1) reasoning tokens dominate the trajectory, and 2) training signals are not uniformly distributed across action tokens. Table~\ref{tab:token-composition} reveals the imbalance: across environments and model scales, action tokens account for less than 16\% of generated tokens in a trajectory. Given this, uniform credit assignment would concentrate most gradient mass on reasoning spans.

\begin{figure*}[t]
  \centering
  \includegraphics[width=1.0\linewidth]{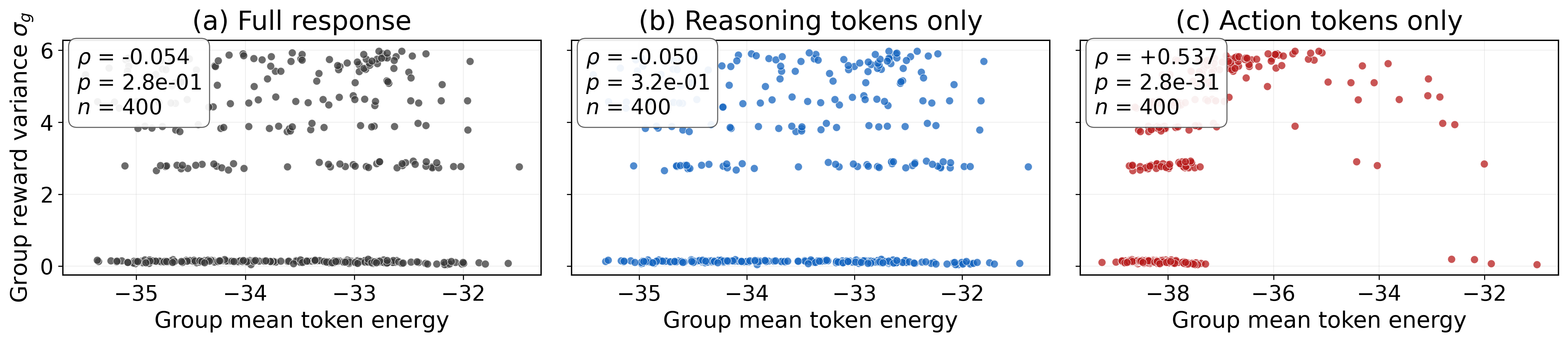}
  \vspace{-5pt}
  \caption{
  \textbf{Detailed energy--reward diagnostic behind Figure~\ref{fig:action-bottleneck}.} We measure the Spearman correlation between mean frozen-reference token energy and group reward variance $\sigma_g$ on Sokoban~3B over three token subsets. Only the action-token energy shows a significant positive correlation ($\rho = +0.537$, $p < 10^{-30}$); full-response and reasoning-only energy are indistinguishable from noise.}
  \label{fig:energy-sigma-corr}
  \vspace{-10pt}
\end{figure*}

To further validate the results in Figure~\ref{fig:action-bottleneck} across different cases, we examine where the training signals live within a trajectory. For each prompt in the training set, we run $G$ independent rollouts under the current policy, each producing a complete multi-turn trajectory and its episodic reward $R(\tau)$. We refer to the $G$ trajectories sampled from the same prompt as a \emph{trajectory group}, with group reward variance $\sigma_g = \mathrm{Var}(R(\tau^{(1)}), \dots, R(\tau^{(G)}))$. A high $\sigma_g$ means the policy sometimes succeeds and sometimes fails on this prompt, so the prompt provides useful learning signals. For each trajectory group, we then compute the mean frozen-reference token energy over three token subsets: the full response, reasoning tokens only, and action tokens only. We measure the Spearman correlation between the mean energy of each subset and $\sigma_g$ across all trajectory groups. Figure~\ref{fig:energy-sigma-corr} shows the result on Sokoban~3B. We find that action-only energy correlates with $\sigma_g$, with $\rho=+0.537$ and $p<10^{-30}$, while full-response and reasoning-only energy are indistinguishable from random noise. This shows that the training signals captured by frozen-reference energy largely stem from action tokens instead of reasoning tokens that dominate in the trajectory.

The observation motivates a token-level reweighting scheme that proceeds in two steps. Specifically, we first use the explicit \texttt{<think>}/\texttt{<answer>} boundary to reallocate gradient mass from reasoning tokens towards environment-facing action tokens. We then use frozen-reference energy within the action span to prioritize positions where the reference model is more uncertain.

\subsection{Reasoning Downweighting and Energy-Based Action Prioritization}
\label{sec:structure-reweight}

For each response token belonging to either a reasoning span $\mathcal{T}_{\mathrm{think}}$ or an action span $\mathcal{T}_{\mathrm{action}}$, as identified by the \texttt{<think>} and \texttt{<answer>} tags, we assign a weight:
\begin{equation}
\label{eq:token-weight}
w_t =
\begin{cases}
\alpha, & t \in \mathcal{T}_{\mathrm{think}}, \\[4pt]
1 + \beta\,\operatorname{sigmoid}\!\left(\dfrac{E_t - \mu_E}{\sigma_E}\right), & t \in \mathcal{T}_{\mathrm{action}},
\end{cases}
\end{equation}
where $\alpha \in [0,1]$ controls how much gradient is retained on reasoning tokens, $\beta \ge 0$ controls how strongly energy modulates the weights on action tokens, and $E_t$ is the frozen-reference token energy from Eq.~\ref{eq:token-energy}. $\mu_E$ and $\sigma_E$ denote the mean and standard deviation of $E_t$ over all action tokens in the current training batch, respectively. The sigmoid function is applied to a batch-normalized action-token energy score, making energy modulation depend on relative uncertainty within the action-token distribution rather than on the raw energy scale.

When setting $\alpha=1$ and $\beta=0$, which gives $w_t=1$ for all tokens, the reweighting mechanism described by Eq.~\ref{eq:token-weight} reduces to standard uniform credit assignment. Our proposed approach uses $\alpha<1$ and $\beta>0$, which simultaneously downweights reasoning tokens and applies energy-based redistribution within the action span. The former simply shifts the gradient mass towards informative action tokens, while the latter additionally prioritizes the action tokens for which the model is more uncertain.
In our experiments, we compare standard \textsc{Uniform} credit assignment against our proposed token reweighting approach, which we refer to as \textsc{ActFocus}.

\subsection{Modified Objective}
\label{sec:objective}

We incorporate the token-level credits by replacing the averaging aggregation in Eq.~\ref{eq:ppo-loss} with a weighted sum:
\begin{equation}
\label{eq:ewaa-loss}
\mathcal{L}(\theta)
=
-\frac{1}{\sum_t w_t}
\sum_t w_t \,
\min\!\Bigl(
\rho_t(\theta)\,\hat{A}_t,\;
\mathrm{clip}\!\bigl(\rho_t(\theta),\,1{-}\epsilon,\,1{+}\epsilon\bigr)\hat{A}_t
\Bigr),
\end{equation}
where $w_t$ is given by Eq.~\ref{eq:token-weight}. 
Tokens with larger $w_t$ contribute more to the policy-gradient update, while tokens with smaller $w_t$ contribute less. Importantly, this modification does not change how the advantage is computed. For PPO, $\hat{A}_t$ still comes from the original critic-based estimator; for GRPO, it still comes from the group-normalized trajectory return.
Thus, the method only changes how the existing advantage signal is distributed across tokens.

We normalize by $\sum_t w_t$ so that changing $\alpha$ or $\beta$ does not simply make the whole loss larger or smaller. Instead, $\alpha$ and $\beta$ control where the gradient mass goes: decreasing $\alpha$ reduces the contribution of reasoning tokens, while increasing $\beta$ gives more weight to high-energy action tokens.

Ignoring PPO clipping for clarity, the weighted objective separates the update into a reasoning component and an action component:
\begin{equation}
\label{eq:grad-decomp}
\nabla_\theta \mathcal{L}
\;=\;
-\frac{1}{\sum_t w_t}\!\left[
\underbrace{\alpha \sum_{t \in \mathcal{T}_{\mathrm{think}}}
\hat{A}_t \nabla_\theta \log \pi_\theta(y_t \mid c_t)}_{\text{reasoning component}}
+
\underbrace{\sum_{t \in \mathcal{T}_{\mathrm{action}}}
w_t \hat{A}_t \nabla_\theta \log \pi_\theta(y_t \mid c_t)}_{\text{action component}}
\right].
\end{equation}

The two terms correspond directly to the two steps introduced at the start of this section. Reducing $\alpha$ reallocates gradient mass from reasoning spans to action spans. Positive $\beta$ then redistributes mass within the action span, giving larger weight to high-energy tokens where the frozen reference model is more uncertain.

\begin{table}[t]
\centering
\caption{PPO final-step results across four environments at two capability levels. Success rate (\%) is reported at step 200 for Sokoban, FrozenLake, and Sudoku, and at step 100 for WebShop. WebShop reports both purchase rate and strict success. Peak values are reported in Table~\ref{tab:main-ppo-grpo-peak}. Best results are \textbf{bolded}.}
\label{tab:main-ppo}
\small
\begin{tabular*}{\textwidth}{@{\extracolsep{\fill}} l ccc cc}
\toprule
\textbf{Method}
  & \textbf{Sokoban}
  & \textbf{FrozenLake}
  & \textbf{Sudoku}
  & \textbf{WebShop\textsubscript{purchase}}
  & \textbf{WebShop\textsubscript{strict}} \\
\midrule
  & 3B & 3B & 1.5B & 3B & 3B \\
Uniform
  & 26.2
  & 16.6
  & 81.2
  & 96.9
  & 22.7 \\
ActFocus
  & \textbf{37.5} {\scriptsize\color{pos}(+11.3)}
  & \textbf{27.0} {\scriptsize\color{pos}(+10.4)}
  & \textbf{95.7} {\scriptsize\color{pos}(+14.5)}
  & \textbf{99.2} {\scriptsize\color{pos}(+2.3)}
  & \textbf{36.7} {\scriptsize\color{pos}(+14.0)} \\
\midrule
  & 0.5B & 0.5B & 0.5B & 1.5B & 1.5B \\
Uniform
  & 12.5
  & 5.1
  & 16.2
  & 94.9
  & 9.8 \\
ActFocus
  & \textbf{34.0} {\scriptsize\color{pos}(+21.5)}
  & \textbf{15.6} {\scriptsize\color{pos}(+10.5)}
  & \textbf{81.4} {\scriptsize\color{pos}(+65.2)}
  & \textbf{98.8} {\scriptsize\color{pos}(+3.9)}
  & \textbf{26.6} {\scriptsize\color{pos}(+16.8)} \\
\bottomrule
\end{tabular*}
\end{table}

\section{Experiments}
\label{sec:experiments}

We evaluate our proposed method on four multi-turn environments spanning distinct reasoning tasks: \textbf{Sokoban}, a spatial planning task where the agent pushes boxes onto targets on a grid; \textbf{FrozenLake}, a navigation task on a stochastic slippery surface; \textbf{Sudoku~4$\times$4}, a constraint-satisfaction task requiring logical deduction; and \textbf{WebShop}~\citep{yao2023webshopscalablerealworldweb}, an e-commerce task involving product search, evaluation, and purchase. For WebShop, we report both purchase rate and strict success: purchase rate measures whether the agent completes a purchase, while strict success additionally requires the purchased item to match the user instruction. We use the \textbf{Qwen2.5-Instruct family} \citep{qwen2025qwen25technicalreport} throughout all experiments. For each environment, we use two model scales to test whether the Action Bottleneck and reweighting benefits persist across different base-model capabilities. We train each model with both PPO~\citep{schulman2017proximalpolicyoptimizationalgorithms} and GRPO~\citep{shao2024deepseekmathpushinglimitsmathematical}, built on veRL following StarPO-S configuration~\citep{wang2025ragenunderstandingselfevolutionllm}. We compare two token-weighting schemes: \textsc{Uniform} applies standard equal credit 
and \textsc{ActFocus} applies the full structure-aware weighting rule in Eq.~\ref{eq:token-weight}. Detailed dataset specifications, rollout settings, and training hyperparameters are provided in Appendix~\ref{app:experiments}.

\subsection{Main Results}
\label{sec:main-results}

All main results in Table~\ref{tab:main-ppo} use a single fixed hyperparameter setting across all four environments and both model scales. We deliberately avoid per-environment tuning to test whether the Action Bottleneck correction generalizes without configuration-specific calibration. Ablations on the choice of $\alpha$ and $\beta$ are deferred to Sec.~\ref{sec:alpha-sweep} and Sec.~\ref{sec:beta-sweep}.

\noindent\textbf{ActFocus improves across all PPO settings.} Table~\ref{tab:main-ppo} shows 
that \textsc{ActFocus} outperforms \textsc{Uniform} in every tested configuration, across four diverse environments and two model scales. The consistent improvement suggests that the gains stem from correcting a shared structural imbalance in agent trajectories rather than fitting to any specific task or model scale. 

\noindent\textbf{ActFocus helps regardless of whether Uniform is learning.} On Sudoku~0.5B, \textsc{ActFocus} raises success from 16.2\% to 81.4\%, showing that credit reweighting can unlock learning when uniform assignment delivers little usable signal. Even when \textsc{Uniform} already learns nontrivial policies, \textsc{ActFocus} still improves performance by 11.3 points on Sokoban~3B, 10.4 points on FrozenLake~3B, and 14.5 points on Sudoku~1.5B. These results suggest that uniform credit suffers from the same structural problem across regimes: it spreads the learning signal over many reasoning tokens, leaving too little credit for the action tokens that determine the outcome.

\noindent\textbf{ActFocus improves decision quality beyond task completion.} On WebShop, \textsc{Uniform} already achieves a high purchase rate, so there is limited room to improve the purchase-rate metric. The more informative signal is strict success, where \textsc{ActFocus} improves from 22.7\% to 36.7\% on WebShop~3B and from 9.8\% to 26.6\% on WebShop~1.5B. The strict success gain shows that the method improves the quality of the selected product, not merely the ability to finish an episode. This matches the role of frozen-reference energy in Sec.~\ref{sec:method}: energy identifies uncertain action positions, and \textsc{ActFocus} assigns them larger credit during training.

\begin{figure*}[t]
  \centering
  \includegraphics[width=1.0\linewidth]{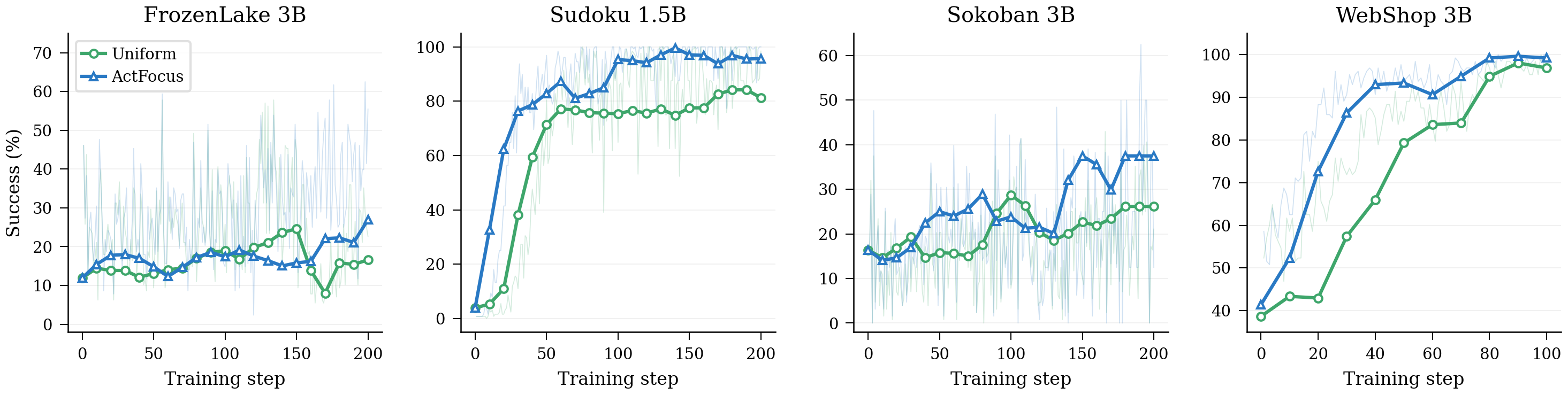}
  \vspace{-5pt}
  \caption{PPO success rate across training steps on FrozenLake~3B, Sudoku~1.5B, Sokoban~3B, and WebShop~3B. Shaded curves denote per-step training success.}
  \label{fig:training-curves}
  \vspace{-10pt}
\end{figure*}

\begin{figure*}[t]
  \centering
  \includegraphics[width=1.0\linewidth]{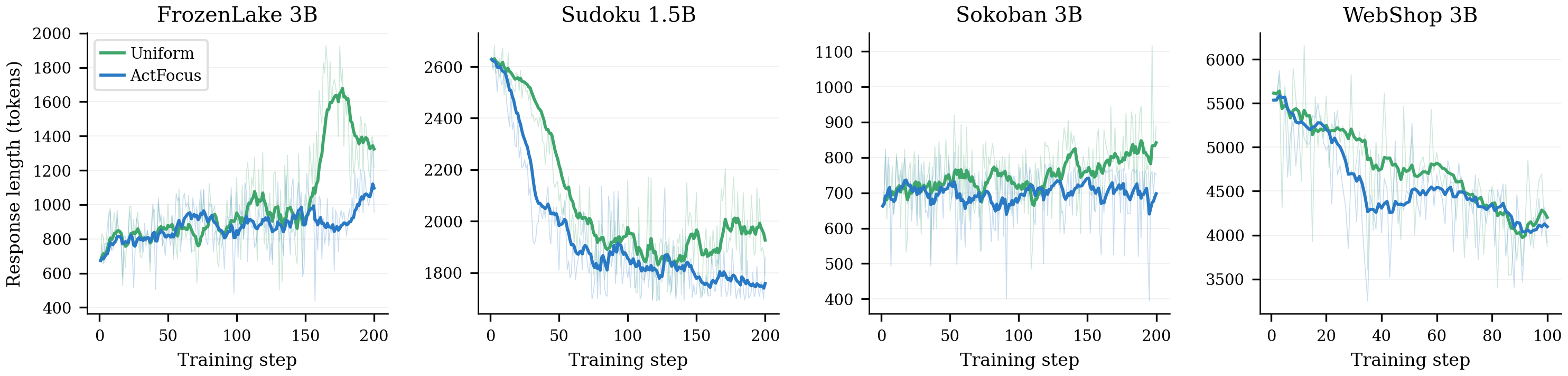}
  \vspace{-5pt}
  \caption{PPO response length across training steps on FrozenLake~3B, Sudoku~1.5B, Sokoban~3B, and WebShop~3B. Shaded curves denote per-step response length.}
  \label{fig:response-length}
  \vspace{-10pt}
\end{figure*}

\subsection{Training Dynamics and Response Efficiency}
\label{sec:training-dynamics}

\noindent\textbf{\textsc{ActFocus} also improves training dynamics.}
Figure~\ref{fig:training-curves} shows that \textsc{ActFocus} improves PPO training dynamics across all four configurations, but the form of improvement differs by environment. On Sudoku~1.5B and WebShop~3B, \textsc{ActFocus} reaches high success earlier than \textsc{Uniform}, indicating faster learning. On Sokoban~3B, the two methods are closer in the early stages, but \textsc{ActFocus} continues to improve in later steps, while \textsc{Uniform} improves more slowly, so the performance gap widens over training. FrozenLake~3B is noisier because of stochastic transitions, but \textsc{ActFocus} still reaches a better final result. Overall, these curves show that the gains in Table~\ref{tab:main-ppo} are reflected in the training process, rather than appearing only at the final checkpoint.

\noindent\textbf{Gains are not caused by longer responses.}
Figure~\ref{fig:response-length} shows that \textsc{ActFocus} produces shorter or more 
stable responses than \textsc{Uniform} across all configurations. On FrozenLake~3B and 
Sokoban~3B, \textsc{Uniform} response length grows throughout training while \textsc{ActFocus} 
keeps it controlled. On Sudoku~1.5B and WebShop~3B, both methods shorten responses over 
training, but \textsc{ActFocus} does so earlier and converges to shorter final lengths. 
Thus, \textsc{ActFocus} improves success without relying on longer reasoning traces, as better credit assignment directs the policy toward action tokens without requiring excessive verbalization.

\subsection{Extension to GRPO}
\label{sec:grpo-extension}
We also evaluate whether the same token-weighting rule transfers to GRPO. Under GRPO, the critic is removed, and a single trajectory-level advantage is assigned to every token in the same rollout, making token weights the primary source of within-trajectory credit 
differentiation. This makes GRPO especially sensitive to the Action Bottleneck: under uniform credit, reasoning tokens dominate the update even though outcomes are mostly determined by sparse action decisions.

\begin{figure*}[t]
  \centering
  \includegraphics[width=1.0\linewidth]{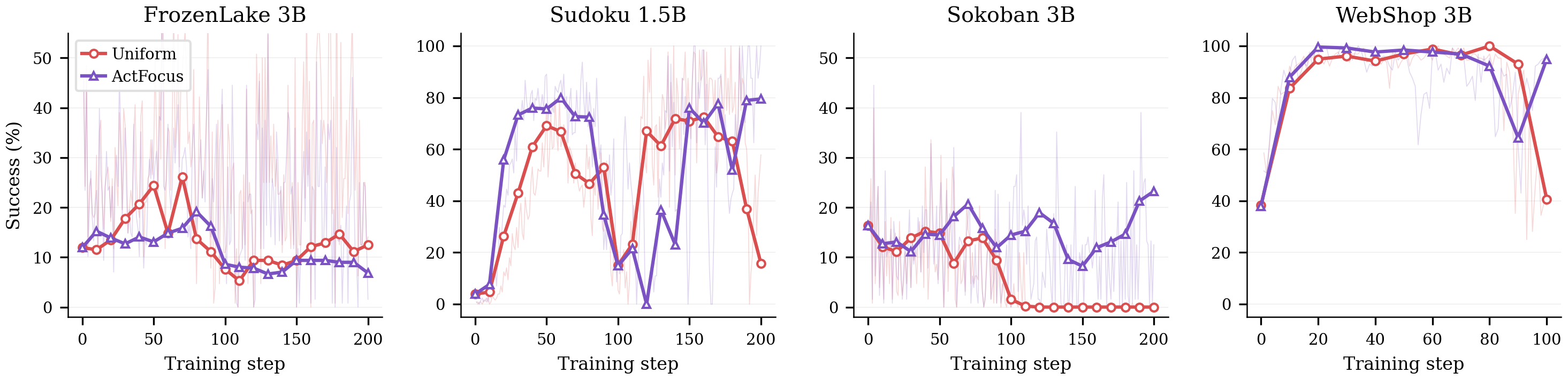}
  \vspace{-5pt}
  \caption{GRPO success rate across training steps on FrozenLake~3B, Sudoku~1.5B, Sokoban~3B, and WebShop~3B. Shaded curves denote per-step training success.}
  \label{fig:grpo-stability}
  \vspace{-10pt}
\end{figure*}

Figure~\ref{fig:grpo-stability} shows that this mismatch often appears as peak-to-final degradation. Across Sokoban~3B, Sudoku~1.5B, and WebShop~3B, \textsc{Uniform} often reaches a useful intermediate policy but fails to preserve it, leading to sharp peak-to-final degradation. \textsc{ActFocus} either prevents the late-stage drop or helps the policy recover from it.

\begin{figure*}[t]
  \centering
  \includegraphics[width=1.0\linewidth]{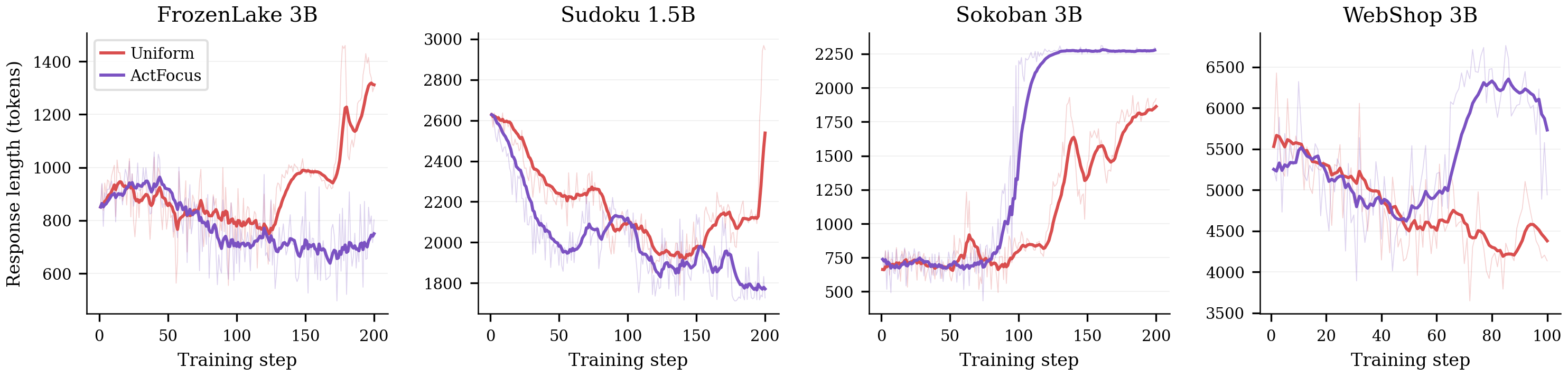}
  \vspace{-5pt}
\caption{GRPO response length across training steps on FrozenLake~3B, Sudoku~1.5B, Sokoban~3B, and WebShop~3B. Shaded curves denote per-step response length.}
  \label{fig:grpo-length}
  \vspace{-10pt}
\end{figure*}

Figure~\ref{fig:grpo-length} provides a complementary view of stabilization. Under GRPO, response length tracks training stability rather than response efficiency. Noisy length curves coincide with unstable policy behavior, while smooth curves indicate that training has settled into a consistent regime, regardless of whether length is rising, falling, or stable. Stability here does not require shorter responses.

\subsection{How Much Reasoning Credit Should Be Retained?}
\label{sec:alpha-sweep}

Although \textsc{ActFocus} improves over \textsc{Uniform} across all main PPO experiments, the optimal $\alpha$ varies by configuration. The intuition is simple: when the base model already reasons well enough for the task, a smaller $\alpha$ removes redundant reasoning updates and concentrates RL on action decisions; when reasoning still needs refinement, a larger $\alpha$ preserves that gradient signal.

\begin{figure}[t]
  \centering
  \includegraphics[width=\textwidth]{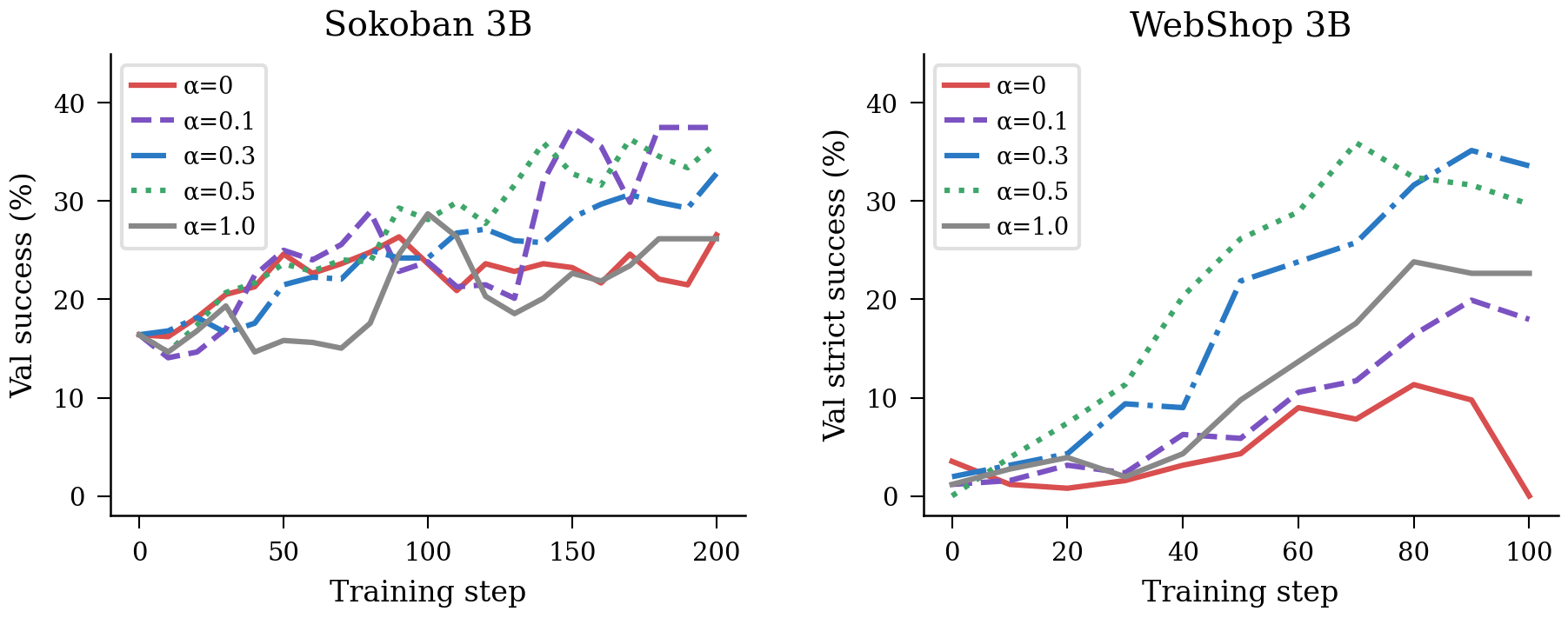}
  \caption{\textbf{Effect of $\alpha$ on training dynamics.}}
  \label{fig:alpha-ablation}
\end{figure}

Figure~\ref{fig:alpha-ablation} confirms this on two representative environments. On 
Sokoban~3B, the best result comes from the aggressive setting $\alpha=0.1$, consistent with a model that already has sufficient spatial reasoning and benefits from focusing updates on action tokens. On WebShop~3B, the optimum shifts to $\alpha=0.3$, and the extreme $\alpha=0$ eventually collapses. Unlike Sokoban, WebShop requires ongoing reasoning refinement, such as query reformulation, evidence aggregation, and product comparison, making full reasoning suppression too aggressive.

The two extremes illustrate why $\alpha$ requires calibration. At $\alpha=1.0$, reasoning tokens retain full credit, so energy-based amplification alone cannot overcome the dominance of the reasoning majority. At $\alpha=0$, reasoning is suppressed entirely, discarding the gradient signal needed for coherent reasoning. The best $\alpha$ sits between these extremes, balancing action focus with reasoning retention.
\vspace{-10pt}
\subsection{How Strongly Should Energy Modulate Action Credit?}
\label{sec:beta-sweep}

\begin{figure}[t]
  \centering

  \begin{subfigure}[t]{0.48\textwidth}
    \centering
    \includegraphics[width=\linewidth]{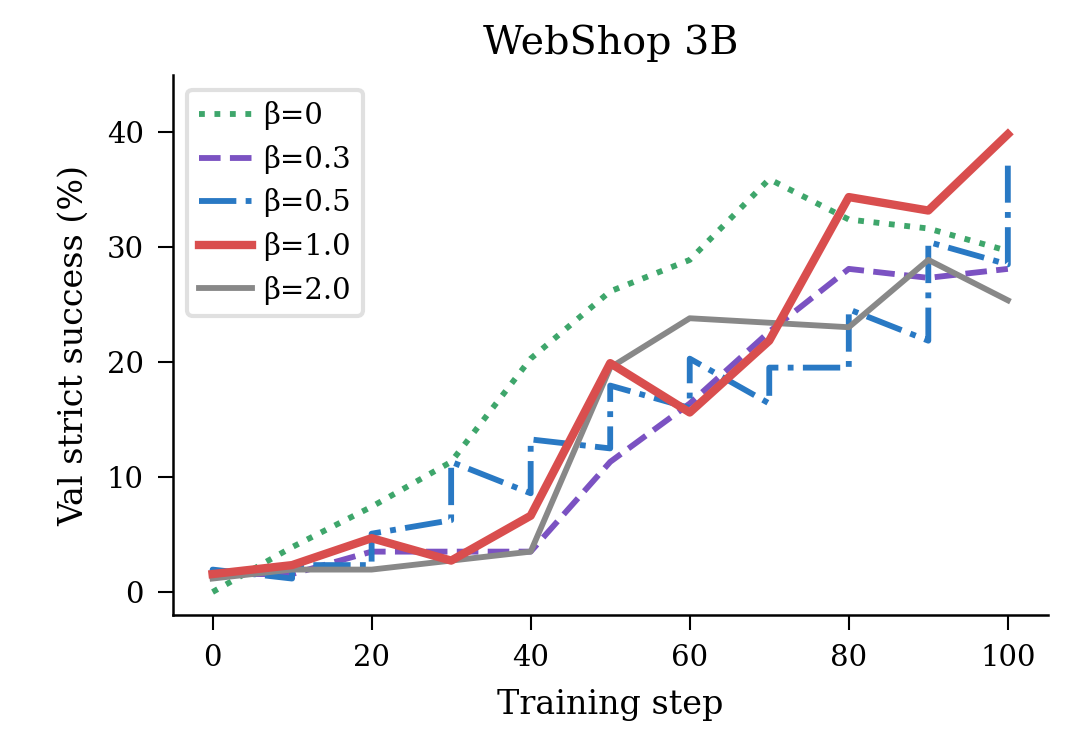}
    \caption{\textbf{Effect of $\beta$.} Sweeping $\beta$ on WebShop~3B: strict success peaks at $\beta{=}1.0$ (39.8\%) and degrades at both extremes.}
    \label{fig:beta-ablation}
  \end{subfigure}
  \hfill
  \begin{subfigure}[t]{0.48\textwidth}
    \centering
    \includegraphics[width=\linewidth]{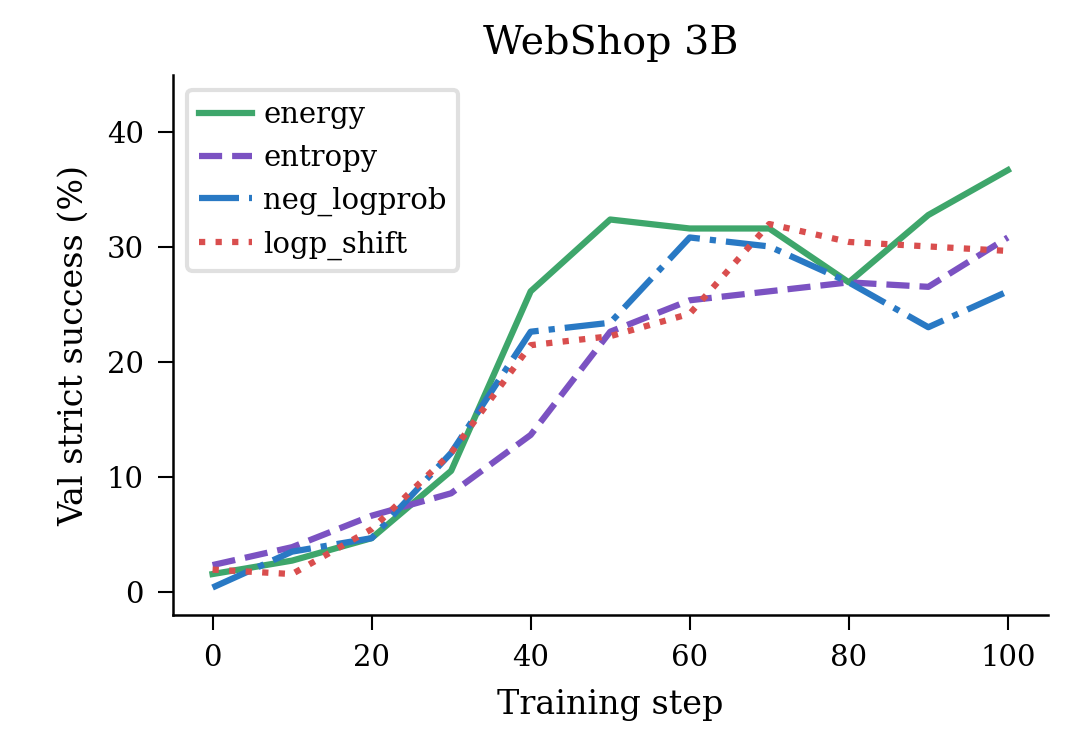}
    \caption{\textbf{Signal comparison.} Replacing energy with alternative signals: energy outperforms entropy, log-likelihood difference, and policy NLL.}
    \label{fig:signal-ablation}
  \end{subfigure}

  \caption{Ablation studies on the reward-shaping design.}
  \label{fig:ablation-side-by-side}
\end{figure}

With the reasoning-retention parameter $\alpha$ fixed, $\beta$ controls how strongly energy reshapes credit within the action span. Setting $\beta=0$ recovers uniform weighting across action tokens, so the action component gain depends only on $\alpha$, while a larger $\beta$ progressively concentrates updates on high-energy positions. We sweep $\beta \in \{0, 0.3, 0.5, 1.0, 2.0\}$ on WebShop~3B and train each setting with PPO for 100 steps. As shown in Figure~\ref{fig:beta-ablation}, performance is non-monotonic. Strict success improves from 29.7\% at $\beta=0$ to 39.8\% at $\beta=1.0$, but drops to 25.4\% at $\beta=2.0$, falling below the $\beta=0$ baseline. This indicates that energy modulation must be strong enough to differentiate uncertain action tokens from confident ones, since treating all action tokens equally fails to exploit the signal energy carries about which decisions are most informative. At the same time, $\beta$ cannot be too large: when the sigmoid-based weighting becomes too peaked, gradient mass collapses onto a small number of high-energy outliers and leaves the remaining action tokens under-trained. 
\vspace{-5pt}

\subsection{Why Energy? A Comparison Against Alternative Signals}
\label{sec:signal-sweep}

A natural question is whether the gains from \textsc{ActFocus} are specific to
energy or an artifact of the reweighting architecture. To test this, we replace
energy $E_t$ with three alternative signals while keeping all other parameters
fixed on WebShop~3B: entropy, policy NLL, and the log-likelihood difference. Figure~\ref{fig:signal-ablation} shows that energy reaches 36.7\% strict success, outperforming entropy (30.9\%), log-likelihood difference (29.4\%), and policy NLL (26.2\%). The gap traces back to the properties motivated in Sec.~\ref{sec:method}: policy-derived signals such as policy NLL and log-likelihood difference shift as the policy updates, making the weighting target a moving one during training, while frozen-reference energy remains stable and retains magnitude information that the frozen-reference entropy discards after normalization. A detailed comparison is included in Appendix~\ref{app:comparison}.

\vspace{-10pt}
\section{Conclusion}
\vspace{-10pt}
In this work, we identify the \emph{Action Bottleneck} as a critical failure mode in agentic reinforcement learning. To address this, we propose a simple structure-aware token reweighting scheme with energy-based redistribution, which redirects gradient mass toward action-relevant decisions. Results across four environments show that this intervention improves learning effectiveness and training stability. Our findings highlight the Action Bottleneck as a central obstacle in agentic RL and show that correcting it provides a principled pathway toward more effective training. Since \textsc{ActFocus} delivers these gains at comparable computational cost to standard PPO and GRPO, it serves as a practical replacement for the uniform token-averaged objective in existing agentic RL pipelines.

\bibliographystyle{plainnat} 
\bibliography{reference}

\newpage
\appendix

\section{Notation}
\label{app:notation}

Table~\ref{tab:notation} summarizes the notation used throughout the paper. Symbols are grouped by category for ease of reference.

\begin{table}[h]
\centering
\caption{Summary of notation used in the paper.}
\label{tab:notation}
\small
\begin{tabular}{ll}
\toprule
\textbf{Symbol} & \textbf{Meaning} \\
\midrule
\multicolumn{2}{l}{\emph{Environment and trajectory}} \\
\midrule
$\mathcal{M} = \langle \mathcal{S}, \mathcal{A}, P, R \rangle$ & Multi-turn agentic MDP \\
$s_k \in \mathcal{S}$ & State at turn $k$ (interaction history) \\
$a_k \in \mathcal{A}$ & Environment-facing action at turn $k$ \\
$o_k$ & Policy response at turn $k$ (reasoning + action) \\
$\tau = (s_1, o_1, \dots, s_K, o_K)$ & Multi-turn trajectory \\
$K$ & Total number of turns in trajectory $\tau$ \\
$T$ & Total number of response tokens in trajectory $\tau$ \\
$R(\tau)$ & Terminal episodic reward \\
$r_t$ & Per-token reward; $r_t = R(\tau)$ if $t = T$, else $0$ \\
\midrule
\multicolumn{2}{l}{\emph{Policy and tokens}} \\
\midrule
$\pi_\theta$ & Trained policy with parameters $\theta$ \\
$\pi_{\theta_{\mathrm{old}}}$ & Rollout policy (snapshot used for sampling) \\
$\pi_{\mathrm{ref}}$ & Frozen reference policy (KL anchor and energy source) \\
$y_t$ & Generic response token at position $t$ \\
$c_t$ & Context preceding $y_t$ (state, prior turns, prior tokens) \\
$\mathcal{T}_{\mathrm{think}}$ & Set of token positions inside \texttt{<think>} spans \\
$\mathcal{T}_{\mathrm{action}}$ & Set of token positions inside \texttt{<answer>} spans \\
\midrule
\multicolumn{2}{l}{\emph{Policy optimization}} \\
\midrule
$J(\theta)$ & Expected episodic return objective \\
$\rho_t(\theta)$ & Token-level importance ratio $\pi_\theta(y_t \mid c_t) / \pi_{\theta_{\mathrm{old}}}(y_t \mid c_t)$ \\
$\hat{A}_t$ & Per-token advantage estimate \\
$\hat{A}_\tau$ & Trajectory-level (GRPO) advantage \\
$V_\phi$ & Token-level value function used by PPO+GAE \\
$\epsilon$ & PPO clipping threshold \\
$(\gamma, \lambda)$ & Discount factor and GAE coefficient \\
$G$ & Number of trajectories per trajectory group (GRPO) \\
$\beta_{\mathrm{KL}}, \beta_{\mathrm{ent}}$ & KL and entropy regularization coefficients \\
\midrule
\multicolumn{2}{l}{\emph{Token energy and reweighting}} \\
\midrule
$f_t^{\mathrm{ref}}$ & Frozen-reference next-token logits at position $t$ \\
$E_t$ & Frozen-reference token energy at position $t$ \\
$\mu_E, \sigma_E$ & Mean and std of $E_t$ over action tokens in a batch \\
$w_t$ & Per-token weight in the structure-aware loss \\
$\alpha$ & Reasoning-retention coefficient ($w_t$ on \texttt{<think>} tokens) \\
$\beta$ & Action component energy modulation strength \\
$\sigma_g$ & Reward variance within a trajectory group \\
\bottomrule
\end{tabular}
\end{table}

\section{Reinforcement Learning Background}
\label{app:rl-objectives}

We follow the notation of Sec.~\ref{sec:prelim}: $y_t$ denotes a generic response token, $c_t$ denotes the context consisting of all tokens preceding $y_t$, including state observations and prior turns within the trajectory, and the token-level importance ratio between the current policy and the rollout policy is:
\begin{equation}
\rho_t(\theta) = \frac{\pi_\theta(y_t \mid c_t)}{\pi_{\theta_{\mathrm{old}}}(y_t \mid c_t)}.
\end{equation}
Throughout this appendix, $T$ denotes the total number of response tokens in a trajectory $\tau$.

\paragraph{General objective.}
The agentic RL objective maximizes expected episodic return over multi-turn trajectories:
\begin{equation}
J(\theta) = \mathbb{E}_{\tau \sim \pi_\theta}\big[R(\tau)\big],
\end{equation}
where $\tau = (s_1, o_1, \dots, s_K, o_K)$ is a $K$-turn trajectory and $R(\tau)$ is the terminal episodic reward defined in Sec.~\ref{sec:prelim}. The corresponding policy gradient is:
\begin{equation}
\nabla_\theta J(\theta) 
= \mathbb{E}_{\tau \sim \pi_\theta}\!\left[\sum_{t=1}^{T} \hat{A}_t\, \nabla_\theta \log \pi_\theta(y_t \mid c_t)\right],
\end{equation}
which distributes credit across all response tokens $y_t$ in $\tau$ via a per-token advantage estimate $\hat{A}_t$. PPO and GRPO differ primarily in how $\hat{A}_t$ is constructed.

\paragraph{Proximal Policy Optimization (PPO).}
PPO~\citep{schulman2017proximalpolicyoptimizationalgorithms} stabilizes the policy update through a clipped surrogate objective combined with a learned token-level value function $V_\phi$, which estimates the expected return-to-go from context $c_t$. The token-level surrogate loss is:
\begin{equation}
\mathcal{L}_{\mathrm{PPO}}(\theta) 
= -\mathbb{E}_t\!\left[\min\!\Big(\rho_t(\theta)\hat{A}_t,\; \clip\!\big(\rho_t(\theta), 1-\epsilon, 1+\epsilon\big)\hat{A}_t\Big)\right],
\end{equation}
where $\epsilon$ is the clipping threshold. The advantage $\hat{A}_t$ is computed via Generalized Advantage Estimation (GAE)~\citep{schulman2018highdimensionalcontinuouscontrolusing}:
\begin{equation}
\hat{A}_t^{\mathrm{GAE}(\gamma, \lambda)} 
= \sum_{l=0}^{T-t} (\gamma\lambda)^l\, \delta_{t+l}, 
\qquad
\delta_t = r_t + \gamma V_\phi(c_{t+1}) - V_\phi(c_t),
\end{equation}
where $\gamma$ and $\lambda$ control the bias and variance of the advantage estimate. Following standard practice in LLM RL, the episodic reward $R(\tau)$ is delivered at the final response token of the trajectory: $r_t = R(\tau)$ if $t = T$, and $r_t = 0$ otherwise. The value function $V_\phi$ is trained jointly with the policy by regressing toward the empirical return-to-go.

\paragraph{Group Relative Policy Optimization (GRPO).}
GRPO~\citep{shao2024deepseekmathpushinglimitsmathematical} removes the value function by sampling a group of $G$ trajectories $\{\tau^{(i)}\}_{i=1}^{G}$ from the same prompt and computing a group-normalized trajectory advantage:
\begin{equation}
\hat{A}_{\tau^{(i)}} = \frac{R(\tau^{(i)}) - \mathrm{mean}\big(\{R(\tau^{(j)})\}_{j=1}^{G}\big)}{\mathrm{std}\big(\{R(\tau^{(j)})\}_{j=1}^{G}\big)},
\end{equation}
which is the explicit group-level form of $\hat{A}_\tau$ in Eq.~\ref{eq:grpo-advantage}. Because this advantage is defined at the trajectory level, it is broadcast to every token within the trajectory: $\hat{A}_t = \hat{A}_{\tau^{(i)}}$ for all tokens $y_t$ in $\tau^{(i)}$. The clipped surrogate then takes the form:
\begin{equation}
\mathcal{L}_{\mathrm{GRPO}}(\theta) 
= -\frac{1}{G}\sum_{i=1}^{G} \mathbb{E}_t\!\left[\min\!\Big(\rho_t(\theta)\hat{A}_{\tau^{(i)}},\; \clip\!\big(\rho_t(\theta), 1-\epsilon, 1+\epsilon\big)\hat{A}_{\tau^{(i)}}\Big)\right].
\end{equation}
GRPO reduces memory and computation by removing the critic, but its uniform broadcasting of $\hat{A}_{\tau}$ across all tokens makes it particularly sensitive to the Action Bottleneck studied in this work.

\paragraph{Auxiliary regularizers.}
Both PPO and GRPO are typically augmented with two regularizers. A KL term against a reference policy $\pi_{\mathrm{ref}}$ controls how far the trained policy drifts from its initialization:
\begin{equation}
\mathcal{L}_{\mathrm{KL}}(\theta) 
= \beta_{\mathrm{KL}}\,\mathbb{E}_t\!\left[D_{\mathrm{KL}}\big(\pi_\theta(\cdot \mid c_t)\,\|\,\pi_{\mathrm{ref}}(\cdot \mid c_t)\big)\right],
\end{equation}
and an entropy bonus $\mathcal{L}_{\mathrm{ent}}(\theta) = -\beta_{\mathrm{ent}}\,\mathbb{E}_t\big[H\big(\pi_\theta(\cdot \mid c_t)\big)\big]$ encourages exploration during early training. The total objective is $\mathcal{L}_{\mathrm{total}} = \mathcal{L}(\theta) + \mathcal{L}_{\mathrm{KL}}(\theta) + \mathcal{L}_{\mathrm{ent}}(\theta)$, where $\mathcal{L}(\theta)$ is the surrogate loss of either PPO or GRPO. In practice, the same frozen base model serves both as the reference policy $\pi_{\mathrm{ref}}$ for KL regularization and as the source of the frozen-reference logits used to compute token energy in Eq.~\ref{eq:token-energy}. The specific values of $\beta_{\mathrm{KL}}$ and $\beta_{\mathrm{ent}}$ used in our experiments are listed in Appendix~\ref{app:experiments}.

\paragraph{Relation to our weighting scheme.}
The structure-aware reweighting in Eq.~\ref{eq:ewaa-loss} is orthogonal to both algorithms' choice of $\hat{A}_t$. Under PPO, $\hat{A}_t$ remains the GAE estimate; under GRPO, it remains the group-normalized trajectory advantage. \textsc{ActFocus} only changes the per-token weight $w_t$ that multiplies the surrogate, so it composes with PPO and GRPO without modifying their advantage estimators or auxiliary regularizers.

\section{Experimental Details}
\label{app:experiments}

\subsection{Task Environment Details}

We evaluate our method on four multi-turn environments that span a broad range of reasoning demands and interaction structures. Sokoban, FrozenLake, and Sudoku are compact symbolic environments with fully specified dynamics, making them well-suited for controlled analysis of reinforcement learning behavior and token-level credit assignment. Although all three are synthetic, they stress different forms of difficulty: irreversible planning in Sokoban, stochastic transitions in FrozenLake, and structured constraint reasoning in Sudoku. To complement these controlled settings, we also include WebShop \citep{yao2023webshopscalablerealworldweb}, a more realistic environment in which the agent must interpret natural-language instructions, navigate a semi-structured interface, and complete a purchase through a sequence of grounded actions. Taken together, these environments allow us to study the Action Bottleneck under both tightly controlled symbolic tasks and a more open-ended web interaction setting.

\paragraph{Sokoban.}
Sokoban is a grid-based planning task in which the agent must push boxes onto target cells within a limited action budget. Unlike standard navigation problems, Sokoban is irreversible: boxes can be pushed but not pulled back, so an early mistake can permanently destroy a valid solution path. This makes the environment a useful testbed for long-horizon planning and action precision in multi-turn agent training.

\paragraph{FrozenLake.}
FrozenLake combines goal-directed planning with stochastic transitions. The agent must navigate a grid world to reach the goal while avoiding holes, but actions are executed on slippery tiles and may lead to unintended outcomes. As a result, good performance depends not only on moving toward the goal but also on choosing actions that remain robust under transition uncertainty.

\paragraph{Sudoku.}
Sudoku evaluates structured reasoning in a sequential decision-making setting. The agent fills a $4 \times 4$ grid one move at a time, subject to row, column, and subgrid constraints. Because each valid action depends on a set of coupled logical conditions, the environment places more emphasis on inference and constraint satisfaction than on exploration or long-horizon control.

\paragraph{WebShop.}
WebShop \citep{yao2023webshopscalablerealworldweb} is a realistic multi-turn web interaction environment in which the agent must satisfy a natural-language shopping request through search, navigation, option selection, and purchase. Compared with the symbolic environments, WebShop introduces richer language grounding, a semi-structured interface, and a less cleanly defined action space. It therefore serves as a complementary setting for studying whether token-level credit assignment mechanisms continue to help when reasoning and action unfold in a more realistic decision-making process.
                                                                                                                
\subsection{Training and Evaluation Settings}

We conduct our experiments using the \textbf{Qwen2.5-Instruct} family \citep{qwen2025qwen25technicalreport} at three scales: 0.5B, 1.5B, and 3B. For each configuration, we train with both PPO~\citep{schulman2017proximalpolicyoptimizationalgorithms} and GRPO~\citep{shao2024deepseekmathpushinglimitsmathematical} under two token-weighting variants: \textsc{Uniform} and \textsc{ActFocus}. All experiments are built on the RAGEN framework~\citep{wang2025ragenunderstandingselfevolutionllm}, which is implemented on top of veRL\footnote{\url{https://github.com/volcengine/verl}}, and are run on a single node with $4\times$ NVIDIA H100 GPUs. We train for 200 rollout--update iterations on Sokoban, FrozenLake, and Sudoku~4$\times$4, and for 100 iterations on WebShop~\citep{yao2023webshopscalablerealworldweb} due to its long-context nature.

Each on-policy update uses grouped rollouts. On Sokoban, FrozenLake, and Sudoku, we use prompt-group size $P=8$ and sample $G=16$ trajectories per prompt; on WebShop, we use $P=16$ and $G=8$. This yields 128 trajectories per training step before filtering. Agents are allowed at most 2 actions per turn and 10 actions per episode on Sokoban and FrozenLake (5-turn horizon), an 8-turn horizon with a 16-step cap on Sudoku, and a 9-turn horizon with a single action per turn on WebShop. Following StarPO-S~\citep{wang2025ragenunderstandingselfevolutionllm}, we apply reward-variance trajectory selection with filter ratio 0.25 on Sokoban and Sudoku. On WebShop and GRPO-based FrozenLake, we use a filter ratio of 1.0 (no filtering), since environmental stochasticity contaminates within-prompt reward variance as a selection signal. PPO uses a mini-batch size of $E=32$, with per-GPU micro-batch sizes scaled to memory (1--8, depending on model scale and environment), under FSDP and Ray for distributed training.

Policy optimization uses GAE with $(\gamma,\lambda)=(1.0,1.0)$, Adam with $(\beta_1,\beta_2)=(0.9,0.999)$, actor learning rate $1\times10^{-6}$, and critic learning rate $1\times10^{-5}$. We use asymmetric clipping $(\epsilon_{\text{low}}, \epsilon_{\text{high}})=(0.2,0.28)$ following~\citet{yu2025dapoopensourcellmreinforcement}, together with entropy regularization of coefficient 0.001. For Sokoban, FrozenLake, and Sudoku, we drop the KL loss term during training and track KL post hoc, following~\citet{yu2025dapoopensourcellmreinforcement}. For WebShop under GRPO, we add a KL loss with coefficient 0.001 and the $k_1$ estimator to reduce late-stage collapse, following the vanilla StarPO setting in~\citet{wang2025ragenunderstandingselfevolutionllm}. We impose a format penalty of $-0.1$ when the agent fails to produce a valid \texttt{<think>}/\texttt{<answer>} structure. Our two token-weighting variants instantiate $w_t$ in Eq.~\ref{eq:token-weight}: \textsc{Uniform} $(\alpha{=}1.0,\beta{=}0)$ and \textsc{ActFocus} $(\alpha{=}0.1,\beta{=}0.5)$. Rollouts are generated with vLLM using \texttt{tensor\_parallel\_size}=1 and \texttt{enforce\_eager}, with a maximum model length of 3600 tokens and 400-token responses for Sokoban, FrozenLake, and Sudoku, and an 8192-token context window for WebShop.

\paragraph{Evaluation.}
We evaluate every 10 training steps and also before training begins (\texttt{val\_before\_train}). For Sokoban, FrozenLake, and Sudoku, we use a fixed evaluation set of 32 distinct prompts rolled out 16 times each under stochastic decoding, for a total of 512 trajectories per evaluation. For WebShop, we use 256 distinct prompts with one rollout each (256 trajectories). Decoding uses a temperature of 0.5 to measure robustness under sampling. Evaluation episodes are truncated using the same per-environment turn and action budgets as in training. We report evaluation success rate on this fixed evaluation set at step 200 for Sokoban, FrozenLake, and Sudoku, and at step 100 for WebShop.

\subsection{Signal Variants for the Energy Ablation}
\label{app:comparison}

This appendix defines the four action-token signals compared in Sec.~\ref{sec:signal-sweep}. 
The goal is to separate the weighting architecture from the choice of signal. 
All variants use the same weighting pipeline. They differ only in the raw score $s_t$ assigned to each action token.

\paragraph{Common pipeline.}
For a batch of action-token positions $\mathcal{T}_{\mathrm{action}}$, we first normalize the raw signal $s_t$ within the action tokens:
\begin{equation}
\mu_s = \frac{1}{|\mathcal{T}_{\mathrm{action}}|}
\sum_{t \in \mathcal{T}_{\mathrm{action}}} s_t,
\qquad
\sigma_s =
\sqrt{
\frac{1}{|\mathcal{T}_{\mathrm{action}}|}
\sum_{t \in \mathcal{T}_{\mathrm{action}}}
(s_t-\mu_s)^2
+\varepsilon
}.
\end{equation}
We then map the normalized score to $(0,1)$:
\begin{equation}
\tilde{s}_t =
\operatorname{sigmoid}\!\left(\frac{s_t-\mu_s}{\sigma_s}\right).
\end{equation}
The final token weight is
\begin{equation}
w_t =
\begin{cases}
\alpha, & t \in \mathcal{T}_{\mathrm{think}}, \\[2pt]
1 + \beta\,\tilde{s}_t, & t \in \mathcal{T}_{\mathrm{action}}.
\end{cases}
\end{equation}
Thus, all four variants use the same $\alpha$, $\beta$, normalization, and sigmoid mapping. 
Only the raw action-token signal $s_t$ changes.

\paragraph{(1) Frozen-reference energy.}
Energy is the signal used by \textsc{ActFocus}. 
It measures uncertainty from the frozen reference model before applying any policy update:
\begin{equation}
s_t^{\mathrm{energy}} = E_t
= -\log \sum_{v \in V} \exp\!\big(f^{\mathrm{ref}}_{t,v}\big),
\end{equation}
where $f^{\mathrm{ref}}_{t,v}$ is the frozen-reference logit for vocabulary token $v$. 
This matches the token-level energy definition in Eq.~\ref{eq:token-energy}. 
Because the reference model is frozen, this score does not move as training progresses.

\paragraph{(2) Frozen-reference entropy.}
Entropy is another fixed-reference uncertainty signal. 
It measures how spread out the frozen reference distribution is:
\begin{equation}
s_t^{\mathrm{entropy}}
=
H\!\big(\pi_{\mathrm{ref}}(\cdot \mid c_t)\big)
=
-\sum_{v \in V} p^{\mathrm{ref}}_{t,v}
\log p^{\mathrm{ref}}_{t,v},
\qquad
p^{\mathrm{ref}}_{t,v}
=
\frac{\exp(f^{\mathrm{ref}}_{t,v})}
{\sum_{u \in V}\exp(f^{\mathrm{ref}}_{t,u})}.
\end{equation}
Entropy uses the same frozen-reference forward pass as energy. The difference is that entropy is computed after softmax normalization, and therefore captures only the dispersion of the distribution, discarding the absolute logit scale.

\paragraph{(3) Policy NLL.}
Policy NLL measures how surprising the sampled token is under the rollout policy:
\begin{equation}
s_t^{\mathrm{nll}}
=
-\log \pi_{\theta_{\mathrm{old}}}(y_t \mid c_t).
\end{equation}
This signal is cheap because PPO already stores rollout log-probabilities. 
However, it is policy-dependent: as the policy changes during training, the meaning and scale of this signal also change.

\paragraph{(4) Log-probability shift.}
Log-probability shift measures how much the rollout policy has moved away from the frozen reference at the sampled token:
\begin{equation}
s_t^{\mathrm{shift}}
=
\log \pi_{\theta_{\mathrm{old}}}(y_t \mid c_t)
-
\log \pi_{\mathrm{ref}}(y_t \mid c_t).
\end{equation}
A large positive value indicates that the rollout policy assigns a higher probability to $y_t$ than the frozen reference does. 
The log-probabilities are computed from logits without additional temperature scaling. 
Unlike energy and entropy, this signal partly moves with the policy because it contains $\pi_{\theta_{\mathrm{old}}}$.

\paragraph{Signal stability.}
Table~\ref{tab:signal-stability} summarizes whether each raw signal changes as the rollout policy updates. The empirical ranking in Sec.~\ref{sec:signal-sweep} matches this stability distinction. Energy and entropy are computed from a fixed model, so they provide stable weighting targets throughout training. Policy NLL and log-probability shift depend on the rollout policy, so their values change as the policy is updated. Among the two frozen-reference signals, energy performs better in our experiments. A useful interpretation is that energy keeps scale information from the reference logits, while entropy keeps only the normalized distribution shape. This makes frozen-reference energy a natural fit for \textsc{ActFocus}: it is stable during training and still preserves confidence information useful for prioritizing uncertain action tokens.

\begin{table}[h]
\centering
\caption{Stability of the four token-level signals during training. Frozen-reference signals provide a fixed weighting target, while policy-derived signals move as the policy changes.}
\label{tab:signal-stability}
\small
\begin{tabular}{lll}
\toprule
\textbf{Signal} & \textbf{Source} & \textbf{Drifts during training?} \\
\midrule
Energy        & Frozen reference $\pi_{\mathrm{ref}}$ & No \\
Entropy       & Frozen reference $\pi_{\mathrm{ref}}$ & No \\
Policy NLL    & Rollout policy $\pi_{\theta_{\mathrm{old}}}$ & Yes \\
Log-probability shift & Rollout policy and frozen reference & Partly \\
\bottomrule
\end{tabular}
\end{table}

\FloatBarrier
\section{Additional Experimental Results}
\label{app:additional}

\begin{table}[!htbp]
\centering
\caption{PPO and GRPO peak performance across four environments. Values are success/purchase rate (\%); deltas are relative to Uniform. WebShop reports both purchase rate and strict success. Best results are \textbf{bolded}.}
\label{tab:main-ppo-grpo-peak}
\small
\begin{tabular*}{\textwidth}{@{\extracolsep{\fill}} ll ccc cc}
\toprule
\textbf{Alg.}
  & \textbf{Method}
  & \textbf{Sokoban}
  & \textbf{FrozenLake}
  & \textbf{Sudoku}
  & \textbf{WebShop\textsubscript{purchase}}
  & \textbf{WebShop\textsubscript{strict}} \\
\midrule

  & & 3B & 3B & 1.5B & 3B & 3B \\

\textsc{PPO}
  & Uniform
  & 28.7
  & 24.6
  & 84.2
  & 98.0
  & 23.8 \\
  & ActFocus
  & \textbf{37.5} {\scriptsize\color{pos}(+8.8)}
  & \textbf{27.0} {\scriptsize\color{pos}(+2.4)}
  & \textbf{99.6} {\scriptsize\color{pos}(+15.4)}
  & \textbf{99.6} {\scriptsize\color{pos}(+1.6)}
  & \textbf{30.5} {\scriptsize\color{pos}(+6.7)} \\

\addlinespace[2pt]

\textsc{GRPO}
  & Uniform
  & 17.8
  & \textbf{26.2}
  & 70.9
  & \textbf{100.0}
  & 44.9 \\
  & ActFocus
  & \textbf{23.2} {\scriptsize\color{pos}(+5.4)}
  & 19.1 {\scriptsize\color{neg}(-7.1)}
  & \textbf{80.9} {\scriptsize\color{pos}(+10.0)}
  & 99.6 {\scriptsize\color{neg}(-0.4)}
  & \textbf{51.6} {\scriptsize\color{pos}(+6.7)} \\

\midrule

  & & 0.5B & 0.5B & 0.5B & 1.5B & 1.5B \\

\textsc{PPO}
  & Uniform
  & \textbf{43.6}
  & 16.2
  & 16.2
  & 94.9
  & 10.9 \\
  & ActFocus
  & 40.4 {\scriptsize\color{neg}(-3.2)}
  & \textbf{20.9} {\scriptsize\color{pos}(+4.7)}
  & \textbf{82.0} {\scriptsize\color{pos}(+65.8)}
  & \textbf{99.2} {\scriptsize\color{pos}(+4.3)}
  & \textbf{27.0} {\scriptsize\color{pos}(+16.1)} \\

\addlinespace[2pt]

\textsc{GRPO}
  & Uniform
  & 20.7
  & 23.8
  & 0.0
  & 99.6
  & 49.2 \\
  & ActFocus
  & \textbf{23.4} {\scriptsize\color{pos}(+2.7)}
  & \textbf{25.2} {\scriptsize\color{pos}(+1.4)}
  & 0.0 {\scriptsize(no learning)}
  & \textbf{100.0} {\scriptsize\color{pos}(+0.4)}
  & \textbf{64.5} {\scriptsize\color{pos}(+15.3)} \\

\bottomrule
\end{tabular*}
\end{table}

\begin{table}[!htbp]
\centering
\caption{GRPO final-step results across four environments at the larger of the two model scales tested per environment. Success rate (\%) is reported at step 200 for Sokoban, FrozenLake, and Sudoku, and at step 100 for WebShop. WebShop reports both purchase rate and strict success. Small-scale GRPO results are omitted because uniform credit largely fails to learn at this scale (e.g., Sudoku~0.5B reaches 0.0 in Table~\ref{tab:main-ppo-grpo-peak}). Uniform values here reflect the peak-to-final degradation under critic-free training (see Sec.~\ref{sec:grpo-extension}). Best results are \textbf{bolded}.}
\label{tab:main-grpo}
\small
\begin{tabular*}{\textwidth}{@{\extracolsep{\fill}} l c c c c c}
\toprule
\textbf{Method}
& \textbf{Sokoban}
& \textbf{FrozenLake}
& \textbf{Sudoku}
& \textbf{WebShop\textsubscript{purchase}}
& \textbf{WebShop\textsubscript{strict}} \\
\midrule

& 3B & 3B & 1.5B & 3B & 3B \\
Uniform
& 0.0
& \textbf{12.5}
& 15.8
& 40.6
& 14.1 \\
ActFocus
& \textbf{23.2} {\scriptsize\color{pos}(+23.2)}
& 6.8 {\scriptsize\color{neg}(-5.7)}
& \textbf{79.5} {\scriptsize\color{pos}(+63.7)}
& \textbf{94.9} {\scriptsize\color{pos}(+54.3)}
& \textbf{51.6} {\scriptsize\color{pos}(+37.5)} \\

\bottomrule
\end{tabular*}
\end{table}

\FloatBarrier

\section{Environment Prompts and Example Rollouts}

\subsection{Prompt Templates}

Across environments, we use the same three-role chat template: a fixed
\texttt{system} message that describes the environment, a per-turn
\texttt{user} message containing the current state, and an
\texttt{assistant} message produced by the policy. The system message
is assembled from the per-environment instruction together with an
optional grid-symbol vocabulary and, when applicable, an enumeration
of admissible actions. The user message follows the same general
structure across environments:
\texttt{(Turn $n$):$\backslash$nState:$\backslash$n\{state\}$\backslash$nYou have
\{actions\_left\} actions left. Always output: <think> [Your thoughts]
</think> <answer> [your answer] </answer> with no extra text. Strictly
follow this format. Max response length: \{max\_tokens\} words (tokens).}

Below, we show, for each of the four environments used in our
experiments, the exact \texttt{system} prompt emitted by the training
pipeline.

\subsubsection{Sokoban (\texttt{SimpleSokoban})}

\begin{tcolorbox}[
    colback=lightgray!10,
    colframe=black,
    title={\textbf{Sokoban System Prompt}},
    breakable
]
\begin{verbatim}
You're a helpful assistant. You are solving the Sokoban puzzle.
You are the player and you need to push all boxes to targets.
When you are right next to a box, you can push it by moving in
the same direction.
You cannot push a box through a wall, and you cannot pull a box.
The answer should be a sequence of actions, like
<answer>Right || Right || Up</answer>

The meaning of each symbol in the state is:
#: wall, _: empty, O: target, √: box on target, X: box, P: player,
S: player on target
Your available actions are:
Up, Down, Left, Right
You can make up to 10 actions, separated by the action separator " || "
\end{verbatim}
\end{tcolorbox}

\subsubsection{Sudoku (\texttt{SimpleSudoku}, 4$\times$4)}

\begin{tcolorbox}[
    colback=lightgray!10,
    colframe=black,
    title={\textbf{Sudoku System Prompt}},
    breakable
]
\begin{verbatim}
You're a helpful assistant. You are solving a 4x4 Sudoku puzzle.
Fill in the grid so that every row, column, and 2x2 box contains
the numbers 1-4 without repetition.
Initial cells are shown in [brackets] and cannot be modified.
Empty cells are shown as dots (.).
Place numbers one at a time using the format:
<answer>place 3 at row 1 col 2</answer> or <answer>1,2,3</answer>
The environment will provide feedback on valid/invalid moves and
show conflicts if any occur.
\end{verbatim}
\end{tcolorbox}

\subsubsection{FrozenLake (\texttt{CoordFrozenLake}, 4$\times$4, slippery)}

\begin{tcolorbox}[
    colback=lightgray!10,
    colframe=black,
    title={\textbf{FrozenLake System Prompt}},
    breakable
]
\begin{verbatim}
You're a helpful assistant. You are solving the FrozenLake puzzle.
The observation includes both a symbol grid and zero-indexed
coordinates for the start, goal, player, and any holes.
Beware that the ice is slippery, so the agent might slide and end
up in an unintended tile.
Respond with a sequence of actions such as
<answer>Left || Up || Up</answer>.

The meaning of each symbol in the state is:
P: player, _: empty, O: hole, G: goal, X: player in hole,
√: player on goal
Your available actions are:
Left, Down, Right, Up
You can make up to 10 actions, separated by the action separator " || "
\end{verbatim}
\end{tcolorbox}

\subsubsection{WebShop}

The WebShop system prompt contains the task description only; the
admissible-action list is inserted into each turn's state string rather
than the system prompt itself.

\begin{tcolorbox}[
    colback=lightgray!10,
    colframe=black,
    title={\textbf{WebShop System Prompt}},
    breakable
]
\begin{verbatim}
You're a helpful assistant. You are browsing an online shop. Based
on the instruction, buy a product that close to the production
description. You need to search, read the search results, pick a
product, choose the size and color and buy. You should only choose
action from the available actions list provided later. Example
process: I need a gingko light and 20x20 pillow cover that is hand
painted. First search[gingko light 20x20 pillow cover hand painted],
answer format: <answer>search[blanket with fleece throw]</answer>.
Valid answer is search[<keywords>] or click[<clickable>].
\end{verbatim}
\end{tcolorbox}

\subsection{Multi-turn State--Action Templates}

Each example below is a verbatim excerpt from an evaluation-time
rollout. We preserve the exact \texttt{system / user / assistant}
role markers, state formatting, and model outputs.

\subsubsection{Sokoban}

\begin{tcolorbox}[
    colback=lightgray!10,
    colframe=black,
    title={\textbf{Sokoban Rollout Example (Turns 1--2)}},
    breakable
]
\textbf{Input:}
\begin{verbatim}
system
You're a helpful assistant. You are solving the Sokoban puzzle.
You are the player and you need to push all boxes to targets.
When you are right next to a box, you can push it by moving in
the same direction.
You cannot push a box through a wall, and you cannot pull a box.
The answer should be a sequence of actions, like
<answer>Right || Right || Up</answer>

The meaning of each symbol in the state is:
#: wall, _: empty, O: target, √: box on target, X: box, P: player,
S: player on target
Your available actions are:
Up, Down, Left, Right
You can make up to 10 actions, separated by the action separator " || "

user

Turn 1:
State:
######
###P_#
###X_#
###__#
###O_#
######
You have 10 actions left. Always output: <think> [Your thoughts]
</think> <answer> [your answer] </answer> with no extra text.
Strictly follow this format. Max response length: 100 words (tokens).
\end{verbatim}

\textbf{Output (Turn 1):}
\begin{verbatim}
<think>The player is right next to a box. The box is not on a target
yet. The target is directly above the player.</think>
<answer>Right || Up</answer>
\end{verbatim}

\textbf{Next User message (Turn 2):}
\begin{verbatim}
user
Reward:
-0.2

Turn 2:
State:
######
###_P#
###X_#
###__#
###O_#
######
You have 8 actions left. Always output: <think> [Your thoughts]
</think> <answer> [your answer] </answer> with no extra text.
Strictly follow this format. Max response length: 100 words (tokens).
\end{verbatim}

\textbf{Output (Turn 2):}
\begin{verbatim}
<think>After pushing the box right, the player is now right next to
the target. The box is now on the target.</think>
<answer>Up</answer>
\end{verbatim}
\end{tcolorbox}

\subsubsection{Sudoku}

\begin{tcolorbox}[
    colback=lightgray!10,
    colframe=black,
    title={\textbf{Sudoku Rollout Example (Turn 1)}},
    breakable
]
\textbf{Input:}
\begin{verbatim}
system
You're a helpful assistant. You are solving a 4x4 Sudoku puzzle.
Fill in the grid so that every row, column, and 2x2 box contains
the numbers 1-4 without repetition.
Initial cells are shown in [brackets] and cannot be modified.
Empty cells are shown as dots (.).
Place numbers one at a time using the format:
<answer>place 3 at row 1 col 2</answer> or <answer>1,2,3</answer>
The environment will provide feedback on valid/invalid moves and
show conflicts if any occur.

user

Turn 1:
State:
==================================================
SUDOKU PUZZLE
==================================================
[3][1]| [2][4]
[4] . |  .  .
-------------
 . [3]| [4] .
 . [4]| [3][2]

Legend: [N]=initial cell, N=user-placed, *N*=conflict, .=empty

VALID NUMBERS FOR EMPTY CELLS:
  - (2,2): [2]
  - (2,3): [1]
  - (2,4): [1, 3]
  - (3,1): [1, 2]
  - (3,4): [1]
  - (4,1): [1]

Progress: 10/16 cells filled (10 initial, 0 placed)
Steps: 0/16
You have 10 actions left. Always output: <think> [Your thoughts]
</think> <answer> [your answer] </answer> with no extra text.
Strictly follow this format. Max response length: 150 words (tokens).
\end{verbatim}

\textbf{Output:}
\begin{verbatim}
<think>Considering the initial placement, the only number that can go
in (2,2) is 2.</think>
<answer>2,2,2</answer>
\end{verbatim}
\end{tcolorbox}

\subsubsection{FrozenLake}

\begin{tcolorbox}[
    colback=lightgray!10,
    colframe=black,
    title={\textbf{FrozenLake Rollout Example (Turns 1--2)}},
    breakable
]
\textbf{Input:}
\begin{verbatim}
system
You're a helpful assistant. You are solving the FrozenLake puzzle.
The observation includes both a symbol grid and zero-indexed
coordinates for the start, goal, player, and any holes.
Beware that the ice is slippery, so the agent might slide and end
up in an unintended tile.
Respond with a sequence of actions such as
<answer>Left || Up || Up</answer>.

The meaning of each symbol in the state is:
P: player, _: empty, O: hole, G: goal, X: player in hole,
√: player on goal
Your available actions are:
Left, Down, Right, Up
You can make up to 10 actions, separated by the action separator " || "

user

Turn 1:
State:
Coordinates:
Board size: 4 rows x 4 cols (zero-indexed).
Start: (2, 3)
Goal: (3, 0)
Player: (2, 3)
Holes: (0, 3), (1, 1)
Grid Map:
___O
_O__
___P
G___
You have 10 actions left. Always output: <think> [Your thoughts]
</think> <answer> [your answer] </answer> with no extra text.
Strictly follow this format. Max response length: 120 words (tokens).
\end{verbatim}

\textbf{Output (Turn 1):}
\begin{verbatim}
<think>The player is at position (2, 3) and needs to move towards
the goal at (3, 0). Moving left is the first logical step.</think>
<answer>Left || Left</answer>
\end{verbatim}

\textbf{Next User message (Turn 2):}
\begin{verbatim}
user
Reward:
0

Turn 2:
State:
Coordinates:
Board size: 4 rows x 4 cols (zero-indexed).
Start: (2, 3)
Goal: (3, 0)
Player: (2, 1)
Holes: (0, 3), (1, 1)
Grid Map:
___O
_O__
_P__
G___
You have 8 actions left. Always output: <think> [Your thoughts]
</think> <answer> [your answer] </answer> with no extra text.
Strictly follow this format. Max response length: 120 words (tokens).
\end{verbatim}

\textbf{Output (Turn 2):}
\begin{verbatim}
<think>After moving left, the player is now at (2, 1). Moving left
again would lead the player into a hole. The next best move is to
move up.</think>
<answer>Up</answer>
\end{verbatim}
\end{tcolorbox}

\subsubsection{WebShop}

\begin{tcolorbox}[
    colback=lightgray!10,
    colframe=black,
    title={\textbf{WebShop Rollout Example (Turn 1)}},
    breakable
]
\textbf{Input:}
\begin{verbatim}
system
You're a helpful assistant. You are browsing an online shop. Based
on the instruction, buy a product that close to the production
description. You need to search, read the search results, pick a
product, choose the size and color and buy. You should only choose
action from the available actions list provided later. Example
process: I need a gingko light and 20x20 pillow cover that is hand
painted. First search[gingko light 20x20 pillow cover hand painted],
answer format: <answer>search[blanket with fleece throw]</answer>.
Valid answer is search[<keywords>] or click[<clickable>].

user

Turn 1:
State:
Instruction: Find me machine wash, wash cold women's fashion hoodies
& sweatshirts for dry clean, tumble dry with color: vintage camo,
and size: large, and price lower than 70.00 dollars.We must buy a
product within 10 actions. It doesn't have to match perfectly with
description.
Search term should not include details like size, color.
Never search for more than 2 times.
Do not be too strict about the description, it's more important to
buy one that is close enough within action limit.
Prioritize click a product in the current page over going to next
page.
Almost never click[next >] for more than 2 times.Almost never
click[< prev] unless you are sure the product is on one of the
previous pages.
If you have less than 3 actions left, just buy the first product you
see in the current page.
If an matching option exists, make sure to click[size] then
click[color], one at a time, before click[buy now], but don't have
to if only 1 action left, in that case you just click[buy now].
Never click description.
You must choose from these actions:search[<content>].
You have 9 actions left. Always output: <think> [Your thoughts]
</think> <answer> [your answer] </answer> with no extra text.
Strictly follow this format. Max response length: 200 words (tokens).
\end{verbatim}

\textbf{Output:}
\begin{verbatim}
<think>Based on the instruction, I need to find a machine washable,
wash cold women's hooded sweatshirt that is vintage camo pattern and
in large size, with a price lower than 70.00 dollars. The search
term should not include size or color.</think>
<answer>search[women hooded sweatshirt vintage camo large machine
wash cold]</answer>
\end{verbatim}
\end{tcolorbox}

\end{document}